\documentclass[
]{ceurart}

\usepackage{listings}
\usepackage{adjustbox}
\usepackage{array}
\usepackage{float}
\begin{document}

\copyrightyear{2024}
\copyrightclause{Copyright for this paper by its authors.
  Use permitted under Creative Commons License Attribution 4.0
  International (CC BY 4.0).}

\conference{CLEF 2024: Conference and Labs of the Evaluation Forum, September 09–12, 2024, Grenoble, France}

\title{Bilingual Sexism Classification: Fine-Tuned XLM-RoBERTa and GPT-3.5 Few-Shot Learning}

\title[mode=sub]{Notebook for the EXIST Lab at CLEF 2024}

\author[1]{AmirMohammad Azadi}[%
email={am_azadi@comp.iust.ac.ir},
]
\cormark[1]
\fnmark[1]
\address[1]{Iran University of Science and Technology, Tehran, Iran}

\author[1]{Baktash Ansari}[%
email={baktash_ansari@comp.iust.ac.ir},
]
\fnmark[1]

\author[1]{Sina Zamani}[%
email={sina_zamani@comp.iust.ac.ir},
]
\fnmark[1]

\author[1]{Sauleh Eetemadi}[%
email={sauleh@iust.ac.ir},
]

\cortext[1]{Corresponding author.}
\fntext[1]{These authors contributed equally.}

\begin{abstract}
  Sexism in online content is a pervasive issue that necessitates effective classification techniques to mitigate its harmful impact. Online platforms often have sexist comments and posts that create a hostile environment, especially for women and minority groups. This content not only spreads harmful stereotypes but also causes emotional harm. Reliable methods are essential to find and remove sexist content, making online spaces safer and more welcoming. Therefore, the sEXism Identification in Social neTworks (EXIST) challenge addresses this issue at CLEF 2024. This study aims to improve sexism identification in bilingual contexts (English and Spanish) by leveraging natural language processing models. The tasks are to determine whether a text is sexist and what the source intention behind it is. We fine-tuned the XLM-RoBERTa model and separately used GPT-3.5 with few-shot learning prompts to classify sexist content. The XLM-RoBERTa model exhibited robust performance in handling complex linguistic structures, while GPT-3.5's few-shot learning capability allowed for rapid adaptation to new data with minimal labeled examples. Our approach using XLM-RoBERTa achieved \textbf{4\textsuperscript{th}} place in the soft-soft evaluation of Task 1 (sexism identification). For Task 2 (source intention), we achieved \textbf{2\textsuperscript{nd}} place in the soft-soft evaluation.
\end{abstract}


\maketitle

\section{Introduction}

Sexism, defined as unfair treatment or prejudice based on a person's sex or gender, is a serious global issue, especially prevalent online where social media can easily spread sexist ideas. Such content harms society, particularly women, by causing emotional distress and promoting gender inequality. Accurate detection and classification of sexist content are essential for making online spaces safer and more inclusive. This research aims to improve the detection and understanding of sexist language online, helping platforms reduce harmful content and promote respectful digital environments. Effective sexism classification supports content moderation and studies on gender-based discrimination in digital communication.

Our study is part of the EXIST 2024 (sEXism Identification in Social Networks) \cite{plaza2024exist, plaza2024extended} shared task, which aims to improve automated sexism detection. Our research focuses on two main tasks: sexism identification and intention detection. Sexism identification involves deciding if a text contains sexist content, while intention detection tries to understand the purpose behind the sexist remarks, categorizing them into the following three types:
\begin{itemize}

    \item 
    "Direct" describes whether the intention was to write a message that is sexist by itself or incites to be sexist.
    
    \item 
    The label "Reported" states that the intention is to report and share a sexist situation suffered by a woman or women in the first or third person.

    \item 
    Lastly, "Judgemental" shows the intention was to judge since the tweet describes sexist situations or behaviors with the aim of condemning them.
    
\end{itemize}

These tasks are crucial for developing systems that can detect and understand sexist language in context.

To address these challenges, we used two techniques in natural language processing: XLM-RoBERTa \cite{conneau2020unsupervised} Fine-Tuning and GPT-3.5 Few-Shot Learning. XLM-RoBERTa, an advanced version of the RoBERTa \cite{liu2019roberta} model, is fine-tuned on the dataset to better recognize and classify sexist content. This method uses the model’s extensive training on a diverse multilingual dataset, making it good at understanding complex language patterns. We also used GPT-3.5 through few-shot learning, which means giving the model a few English and Spanish tweets from the dataset in each prompt to help it adapt to specific tasks. This approach takes advantage of GPT-3.5’s large-scale training and its ability to understand the context and generate annotations with little extra data.

The rest of this paper is organized as follows: In section 2, we describe the datasets used in our study, explaining their structure. Section 3 details our methodology, including the specific setups for XLM-RoBERTa Fine-Tuning and GPT-3.5 Few-Shot Learning. In section 4, we present the results of our experiments, compare the performance of both methods using various measures, and analyze how well they detect and classify sexist content. Finally, we discuss our findings, suggest potential improvements, and outline directions for future research in automated sexism detection in the concluding sections.

\section{Dataset}

To address label bias in the annotation process, which can arise from socio-demographic differences among annotators or subjective labeling, the EXIST campaign considers some demographic parameters including: gender, age, country, study-level, and ethnicity. Each tweet was annotated by six crowdsourcing annotators selected through Prolific, following guidelines from gender experts.

The EXIST 2024 dataset incorporates multiple types of sexist expressions, including descriptive or reported assertions where the sexist message is a description of sexist behavior. In particular, the dataset is composed of more than 10,000 tweets both in English and Spanish, divided into a test set (2,076 tweets), a development set (1,038 tweets), and a training set (6,920 tweets).

For each sample, the following attributes are provided in a JSON format:
\begin{itemize}
    \item 
    {\textbf{"id\_EXIST"}: a unique identifier for the tweet}
    \item 
    {\textbf{"lang"}: the languages of the text (“en” or “es”)}
    \item 
    {\textbf{"tweet"}: the text of the tweet}
    \item 
    {\textbf{"number\_annotators"}: the number of persons that have annotated the tweet}
    \item 
    {\textbf{"annotators"}: a unique identifier for each of the annotators}
    \item 
    {\textbf{"gender\_annotators"}: the gender of the different annotators. Possible values are: “F” and “M”, for female and male respectively}
    \item 
    {\textbf{"age\_annotators"}: the age group of the different annotators. Possible values are: 18-22, 23-45, and 46+}
    \item 
    {\textbf{"ethnicity\_annotators"}: the self-reported ethnicity of the different annotators. Possible values are: “Black or African America”, “Hispano or Latino”, “White or Caucasian”, “Multiracial”, “Asian”, “Asian Indian” and “Middle Eastern”}
    \item 
    {\textbf{"study\_level\_annotators"}: the self-reported level of study achieved by the different annotators. Possible values are: “Less than high school diploma”, “High school degree or equivalent”, “Bachelor’s degree”, “Master’s degree” and “Doctorate”}
    \item 
    {\textbf{"country\_annotators"}: the self-reported country where the different annotators live in}
    \item 
    {\textbf{"labels\_task1"}: a set of labels (one for each of the annotators) that indicate if the tweet contains sexist expressions or refers to sexist behaviors or not. Possible values are: “YES” and “NO”}
    \item 
    {\textbf{"labels\_task2"}: a set of labels (one for each of the annotators) recording the intention of the person who wrote the tweet. Possible labels are: “DIRECT”, “REPORTED”, “JUDGEMENTAL”, “-”, and “UNKNOWN”}
    \item 
    {\textbf{"labels\_task3"}: a set of arrays of labels (one array for each of the annotators) indicating the type or types of sexism that are found in the tweet. Possible labels are: “IDEOLOGICAL- INEQUALITY”, “STEREOTYPING-DOMINANCE”, “OBJECTIFICATION”, “SEXUAL-VIOLENCE”, “MISOGYNY-NON-SEXUAL-VIOLENCE”, “-”, and “UNKNOWN”}
    \item 
    {\textbf{“split”}: subset within the dataset the tweet belongs to (“TRAIN”, “DEV”, “TEST” + “EN”/”ES”)}
\end{itemize}

In sexism identification, natural language expressions often do not have a single, clear interpretation. To address this, the learning with disagreements paradigm allows systems to learn from datasets that include all annotator opinions rather than a single aggregated label. Following this method, we will provide all annotations per instance from six different annotators, capturing the diversity of views. We determined the final label using a majority voting method, ensuring that the most commonly assigned label among annotators represents the classification.

It should be noted that for Tasks 2 and 3, hard labels are assigned exclusively to tweets identified as sexist (label "YES" for Task 1). Tweets not categorized as sexist receive a label of “–”, and those lacking a label from annotators are marked as "UNKNOWN." The test set is composed solely of the following attributes: "id\_EXIST", "lang", "tweet" and "split."

\section{Methodology}

In this study, we employed two distinct methodologies to tackle the challenge of characterizing sexism on social networks. The first approach involved fine-tuning several state-of-the-art transformer models, namely XLM-RoBERTa, mBERT \cite{devlin2019bert}, deBERTa \cite{he2021deberta}, and BERTIN \cite{delarosa2022bertin}, on the provided dataset. The second approach leveraged the few-shot learning capabilities of GPT-3.5. Below, we provide detailed descriptions of each approach.

\subsection{Fine-Tuning Pre-trained Transformer Models}

This section describes adapting transformer models like XLM-RoBERTa and mBERT for sexism detection through hyper-parameter tuning and optimization techniques to improve performance. The models are evaluated using accuracy, precision, recall, and F1-score.

\subsubsection{Model Selection and Fine-Tuning}
We selected several pre-trained transformer models for fine-tuning. The models are as follows:

\begin{itemize}
    \item \textbf{XLM-RoBERTa}: A multilingual variant of RoBERTa, trained on 100 languages, known for robust performance across various multilingual benchmarks
    \item \textbf{mBERT}: Multilingual BERT, trained on Wikipedia pages from 104 languages, capable of processing multiple languages simultaneously
    \item \textbf{deBERTa}: An improved version of BERT with disentangled attention and an enhanced mask decoder, capturing word dependencies more effectively
    \item \textbf{BERTIN}: A Spanish language model based on BERT, fine-tuned on a large corpus of Spanish texts, tailored for Spanish NLP tasks
\end{itemize}

Each model was fine-tuned on the training set using the following steps:

\begin{enumerate}
    \item \textbf{Training Setup}: The models were initialized with pre-trained weights and adapted to our specific task.
    
    \newpage

    \begin{table}[h!]
    \centering
    \caption{Comparison of Different Models}
    \label{tab:table1}
    \begin{adjustbox}{max width=\textwidth}
    \renewcommand{\arraystretch}{1.2}
    \begin{tabular}{|l|c|c|c|c|}
    \hline
    \textbf{Model} & \textbf{Accuracy} & \textbf{Precision} & \textbf{Recall} & \textbf{F1-score} \\
    \hline
    XLM-RoBERTa/raw & 0.83 & 0.83 & 0.83 & 0.83 \\
    XLM-RoBERTa/param tuning & \textbf{0.87} & \textbf{0.87} & \textbf{0.87} & \textbf{0.87} \\
    mBERT & 0.80 & 0.80 & 0.80 & 0.80 \\
    deBERTa & 0.81 & 0.81 & 0.81 & 0.81 \\
    BERTIN & 0.77 & 0.78 & 0.76 & 0.76 \\
    \hline
    \end{tabular}
    \end{adjustbox}
    \end{table}

    \item \textbf{Hyper-Parameter Tuning}: The best performing model, XLM-RoBERTa, as shown in Table \ref{tab:table1}, underwent extensive hyper-parameter tuning. Tuned parameters included the following:
    \begin{itemize}
        \item Learning rate
        \item Weight decay
        \item Number of epochs
    \end{itemize}
    \item \textbf{Optimization and Early Stopping}: We used the AdamW optimizer along with early stopping to prevent overfitting. A learning rate scheduler was employed to adjust the learning rate dynamically during training.
\end{enumerate}

\subsubsection{Model Evaluation and Analysis}
To evaluate the model, we used the validation set to monitor performance metrics such as accuracy, precision, recall, and F1-score. Additionally, we analyzed mislabeled tweets to understand the sources of error and identify patterns that could inform further improvements.

\subsubsection{Label Extraction}
We generated two types of labels for output including the following:
\begin{itemize}
    \item \textbf{Hard Labels}: Direct output from the model indicating the predicted class
    \item \textbf{Soft Labels}: Probabilities for each class obtained by applying the softmax function to the last layer's output. This was calculated by extracting the logits from the final layer and normalizing them
\end{itemize}

\subsection{Few-Shot Learning with GPT-3.5}

This section explains using GPT-3.5 for sexism classification with few-shot learning, leveraging minimal data for training. It focuses on prompt design and evaluation metrics like accuracy, tailored for handling multilingual input.

\subsubsection{Prompt Design}
We employed few-shot learning with GPT-3.5, leveraging its ability to understand context with minimal training examples. For each prompt, we randomly selected 3 English and 3 Spanish tweets from the training dataset, including the annotator votes to incorporate the learning with disagreement method.

\subsubsection{Model Execution}
Given the constraints of GPT-3.5 in providing probability scores, we only extracted hard labels from its outputs. The prompts were designed to include the following:
\begin{itemize}
    \item The tweet text
    \item Annotator votes, highlighting the disagreement and consensus among human annotators
    \item A clear task description asking GPT-3.5 to classify the tweet
\end{itemize}

\subsubsection{Evaluation}
We assessed GPT-3.5’s performance using the same metrics as for the transformer models. Given the nature of few-shot learning, the evaluation was primarily focused on accuracy and the ability of the model to handle multilingual input with minimal examples.

\section{Results}

In this section, we present the results of our sexism detection methodologies on social networks. We evaluated the performance using various metrics, including ICM-Soft, ICM-Soft Norm, Cross Entropy for soft labels, and ICM-Hard, ICM-Hard Norm, and F1 for hard labels. The tables below summarize the performance across all data, English tweets, and Spanish tweets. The baselines used for comparison are as follows:
\begin{itemize}
    \item \textbf{EXIST2024-test\_gold}: since the ICM measure is unbounded, a baseline that perfectly predicts the ground truth is considered to provide the best possible reference.
    \item \textbf{EXIST2024-test\_majority-class}: non-informative baseline that classifies all instances as the majority class
    \item \textbf{EXIST2024-test\_minority-class}: non-informative baseline that classifies all instances as the minority class
\end{itemize}

For all tasks and evaluation types (hard-hard and soft-soft), the official metric used is the Information Contrast Measure (ICM). ICM is a similarity function that generalizes Pointwise Mutual Information (PMI) and evaluates system outputs in classification problems by computing their similarity to the ground truth categories \cite{amigo-delgado-2022-evaluating}.

\subsection{Task 1}

This task involves determining whether a given tweet contains sexist content, evaluated through various metrics. The tables present performance metrics for different models on overall data, English tweets, and Spanish tweets. Metrics measure the accuracy and similarity of the models' predictions to the ground truth. The following tables are the results for task 1 in three categories containing overall result in table \ref{tab:table2}, English tweets in table \ref{tab:table3}, and Spanish tweets in table \ref{tab:table4}.

\begin{table}[h!]
\centering
\caption{Task 1: Overall Results}
\label{tab:table2}
\begin{adjustbox}{max width=\textwidth}
\renewcommand{\arraystretch}{1.2}
\begin{tabular}{|l|c|c|c|c|c|c|}
\hline
\textbf{Model} & \textbf{ICM-Soft} & \textbf{ICM-Soft Norm} & \textbf{Cross Entropy} & \textbf{ICM-Hard} & \textbf{ICM-Hard Norm} & \textbf{F1} \\
\hline
EXIST2024-test\_gold & 3.12 & 1.00 & 0.55 & 0.99 & 1.00 & 1.00 \\
EXIST2024-test\_majority-class & -2.36 & 0.12 & 4.61 & -0.44 & 0.28 & 0.00 \\
EXIST2024-test\_minority-class & -3.07 & 0.01 & 5.36 & -0.57 & 0.21 & 0.57 \\
\hline
XLM-RoBERTa (BAZI\_1) & \textbf{0.82} & \textbf{0.63} & \textbf{0.98} & \textbf{0.55} & \textbf{0.78} & \textbf{0.78} \\
GPT-3.5 (BAZI\_2) & - & - & - & 0.35 & 0.67 & 0.71 \\
\hline
\end{tabular}
\end{adjustbox}
\end{table}

\begin{table}[h!]
\centering
\caption{Task 1: English Tweets}
\label{tab:table3}
\begin{adjustbox}{max width=\textwidth}
\renewcommand{\arraystretch}{1.2}
\begin{tabular}{|l|c|c|c|c|c|c|}
\hline
\textbf{Model} & \textbf{ICM-Soft} & \textbf{ICM-Soft Norm} & \textbf{Cross Entropy} & \textbf{ICM-Hard} & \textbf{ICM-Hard Norm} & \textbf{F1} \\
\hline
EXIST2024-test\_gold & 3.11 & 1.00 & 0.58 & 0.98 & 1.00 & 1.00 \\
EXIST2024-test\_majority-class & -2.20 & 0.15 & 4.22 & -0.40 & 0.30 & 0.00 \\
EXIST2024-test\_minority-class & -3.82 & 0.00 & 5.57 & -0.66 & 0.16 & 0.53 \\
\hline
XLM-RoBERTa (BAZI\_1) & \textbf{0.66} & \textbf{0.60} & \textbf{1.02} & \textbf{0.55} & \textbf{0.78} & \textbf{0.75} \\
GPT-3.5 (BAZI\_2) & - & - & - & 0.37 & 0.69 & 0.70 \\
\hline
\end{tabular}
\end{adjustbox}
\end{table}

\begin{table}[h!]
\centering
\caption{Task 1: Spanish Tweets}
\label{tab:table4}
\begin{adjustbox}{max width=\textwidth}
\renewcommand{\arraystretch}{1.2}
\begin{tabular}{|l|c|c|c|c|c|c|}
\hline
\textbf{Model} & \textbf{ICM-Soft} & \textbf{ICM-Soft Norm} & \textbf{Cross Entropy} & \textbf{ICM-Hard} & \textbf{ICM-Hard Norm} & \textbf{F1} \\
\hline
EXIST2024-test\_gold & 3.12 & 1.00 & 0.52 & 1.00 & 1.00 & 1.00 \\
EXIST2024-test\_majority-class & -2.54 & 0.09 & 4.96 & -0.49 & 0.26 & 0.00 \\
EXIST2024-test\_minority-class & -2.57 & 0.09 & 5.00 & -0.51 & 0.24 & 0.60 \\
\hline
XLM-RoBERTa (BAZI\_1) & \textbf{0.90} & \textbf{0.64} & \textbf{0.93} & \textbf{0.53} & \textbf{0.77} & \textbf{0.79} \\
GPT-3.5 (BAZI\_2) & - & - & - & 0.31 & 0.66 & 0.72 \\
\hline
\end{tabular}
\end{adjustbox}
\end{table}

\newpage

\subsection{Task 2}

This task aims to determine the intention behind sexist remarks in tweets. The tables show the performance of different models on overall data, English tweets, and Spanish tweets, using the same metrics as in Task 1. These metrics assess how well the models can categorize tweets based on the perceived intention, such as whether the remark is direct, reported, or judgmental. The following tables are the results for task 2 in three categories containing overall result in table \ref{tab:table5}, English tweets in table \ref{tab:table6}, and Spanish tweets in table \ref{tab:table7}.

\begin{table}[h!]
\centering
\caption{Task 2: Overall Results}
\label{tab:table5}
\begin{adjustbox}{max width=\textwidth}
\renewcommand{\arraystretch}{1.2}
\begin{tabular}{|l|c|c|c|c|c|c|}
\hline
\textbf{Model} & \textbf{ICM-Soft} & \textbf{ICM-Soft Norm} & \textbf{Cross Entropy} & \textbf{ICM-Hard} & \textbf{ICM-Hard Norm} & \textbf{F1} \\
\hline
EXIST2024-test\_gold & 6.20 & 1.00 & 0.91 & 1.54 & 1.00 & 1.00 \\
EXIST2024-test\_majority-class & -0.54 & 0.06 & 4.62 & -0.95 & 0.19 & 0.16 \\
EXIST2024-test\_minority-class & -32.96 & 0.00 & 8.85 & -3.15 & 0.00 & 0.03 \\
\hline
XLM-RoBERTa (BAZI\_1) & \textbf{-1.35} & \textbf{0.39} & \textbf{1.78} & \textbf{0.19} & \textbf{0.56} & \textbf{0.48} \\
GPT-3.5 (BAZI\_2) & - & - & - & 0.04 & 0.51 & 0.43 \\
\hline
\end{tabular}
\end{adjustbox}
\end{table}

\begin{table}[h!]
\centering
\caption{Task 2: English Tweets}
\label{tab:table6}
\begin{adjustbox}{max width=\textwidth}
\renewcommand{\arraystretch}{1.2}
\begin{tabular}{|l|c|c|c|c|c|c|}
\hline
\textbf{Model} & \textbf{ICM-Soft} & \textbf{ICM-Soft Norm} & \textbf{Cross Entropy} & \textbf{ICM-Hard} & \textbf{ICM-Hard Norm} & \textbf{F1} \\
\hline
EXIST2024-test\_gold & 6.12 & 1.00 & 0.94 & 1.44 & 1.00 & 1.00 \\
EXIST2024-test\_majority-class & -5.20 & 0.07 & 4.23 & -0.85 & 0.20 & 0.16 \\
EXIST2024-test\_minority-class & -39.49 & 0.00 & 8.96 & -3.47 & 0.00 & 0.02 \\
\hline
XLM-RoBERTa (BAZI\_1) & \textbf{-1.68} & \textbf{0.36} & \textbf{1.82} & 0.06 & 0.52 & \textbf{0.43} \\
GPT-3.5 (BAZI\_2) & - & - & - & \textbf{0.09} & \textbf{0.53} & 0.40 \\
\hline
\end{tabular}
\end{adjustbox}
\end{table}

\begin{table}[h!]
\centering
\caption{Task 2: Spanish Tweets}
\label{tab:table7}
\begin{adjustbox}{max width=\textwidth}
\renewcommand{\arraystretch}{1.2}
\begin{tabular}{|l|c|c|c|c|c|c|}
\hline
\textbf{Model} & \textbf{ICM-Soft} & \textbf{ICM-Soft Norm} & \textbf{Cross Entropy} & \textbf{ICM-Hard} & \textbf{ICM-Hard Norm} & \textbf{F1} \\
\hline
EXIST2024-test\_gold & 6.24 & 1.00 & 0.89 & 1.60 & 1.00 & 1.00 \\
EXIST2024-test\_majority-class & -5.67 & 0.05 & 4.97 & -1.04 & 0.18 & 0.15 \\
EXIST2024-test\_minority-class & -28.71 & 0.00 & 8.76 & -2.94 & 0.00 & 0.03 \\
\hline
XLM-RoBERTa (BAZI\_1) & \textbf{-1.15} & \textbf{0.40} & \textbf{1.75} & \textbf{0.27} & \textbf{0.59} & \textbf{0.52} \\
GPT-3.5 (BAZI\_2) & - & - & - & -0.02 & 0.49 & 0.44 \\
\hline
\end{tabular}
\end{adjustbox}
\end{table}

\section{Conclusion}

In this study, we tackled the challenge of detecting and classifying sexist content in bilingual contexts (English and Spanish) using advanced natural language processing techniques. We fine-tuned the XLM-RoBERTa model and leveraged GPT-3.5 for few-shot learning to address the EXIST 2024 shared tasks. Our results demonstrated the robustness of the XLM-RoBERTa model in handling complex linguistic structures and the adaptability of GPT-3.5 with minimal labeled examples. Specifically, our XLM-RoBERTa model achieved \textbf{4\textsuperscript{th}} place in the soft-soft evaluation of Task 1 (sexism identification) and \textbf{2\textsuperscript{nd}} place in the soft-soft evaluation of Task 2 (source intention). These results highlight the effectiveness of transformer-based models and few-shot learning in addressing the nuances of sexist language in social media content.

\section{Future Work}

While our approaches yielded promising results, several areas for future improvement and research have been identified. These suggestions are as follows:

\begin{itemize}
    \item \textbf{Enhanced Few-Shot Prompting}: For few-shot prompting, instead of selecting samples randomly, we plan to export the embeddings of the samples from our fine-tuned XLM-RoBERTa model. Using cosine similarity, we will identify the most similar and least similar samples from the training set for each sample in the test set. Although we have found 10 most similar and 10 least similar samples for each test sample, we did not have sufficient time to make inferences. This method could potentially improve the performance of GPT-3.5 in few-shot learning scenarios.

    \item \textbf{Data Augmentation}: We aim to gather additional sexist data for future experiments. Data augmentation techniques, such as replacing synonyms of certain words or translating English tweets to Spanish and vice versa, can be employed to enhance the dataset's diversity and robustness.

    \item \textbf{Fine-Tuning Stronger Models}: In future work, we plan to fine-tune stronger models than XLM-RoBERTa, such as newer versions of transformer models or other advanced architectures, to further boost performance in sexism detection and classification tasks.

    \item \textbf{Incorporating Demographic Annotators' Information}: We aim to use the demographic information about annotators that is provided in the dataset. This includes details such as gender, age, ethnicity, and education level. Incorporating these demographic attributes can provide several advantages as:
    \begin{itemize}
        \item \textbf{Bias Mitigation}: Understanding the demographic background of annotators can help identify and mitigate biases in the annotations, leading to fairer and more balanced models.
        \item \textbf{Enhanced Model Performance}: Demographic information can provide additional context that may improve the model's understanding of nuanced language use and cultural differences, thereby enhancing its classification accuracy.
        \item \textbf{Richer Insights}: Including demographic data allows for a more detailed analysis of how different groups perceive and annotate sexist content, contributing to more comprehensive insights into sexism detection.
    \end{itemize}

\end{itemize}

By addressing these areas, we aim to further refine our methodologies and contribute to the development of more effective and robust systems for automated sexism detection in online content.

\bibliography{sample-ceur}

\end{document}